\documentclass[10pt]{article}
\usepackage[utf8]{inputenc}
\usepackage{authblk}
\usepackage{biblatex}
\usepackage{xcolor}
\usepackage{subcaption}
\usepackage{enumitem}
\usepackage{biblatex}
\usepackage{hyperref}
\usepackage{xurl}
\hypersetup{breaklinks=true}
\usepackage{graphicx}
\graphicspath{Images/}
\usepackage{subfiles} 

\title{Stronger Baseline Models -- A Key Requirement for Aligning Machine Learning Research with Clinical Utility}
\date{}

\author[1-2]{Nathan Wolfrath}
\author[4]{Joel Wolfrath}
\author[1]{Hengrui Hu}
\author[1]{Anjishnu Banerjee}
\author[1-3]{Anai N. Kothari}

\affil[ ]{\small}

\affil[1]{Data Science Institute, Medical College of Wisconsin}
\affil[2]{Bud and Sue Selig Hub for Surgical Data Science, Medical College of Wisconsin}
\affil[3]{Department of Surgery, Medical College of Wisconsin}
\affil[4]{Department of Computer Science and Engineering, University of Minnesota}

\affil[ ]{\newline *Corresponding author: akothari@mcw.edu}
\affil[ ]{Contributing authors: \{nwolfrath, hhu, abanerjee\}@mcw.edu, wolfr046@umn.edu}

\begin{document}
\maketitle

\begin{abstract}
Machine Learning (ML) research has increased substantially in recent years, due to the success of predictive modeling across diverse application domains.
However, well-known barriers exist when attempting to deploy ML models in high-stakes, clinical settings, including lack of model transparency (or the inability to audit the inference process), large training data requirements with siloed data sources, and complicated metrics for measuring model utility.
In this work, we show empirically that including \textit{stronger baseline models} in healthcare ML evaluations has important downstream effects that aid practitioners in addressing these challenges.
Through a series of case studies, we find that the common practice of omitting baselines or comparing against a weak baseline model (e.g. a linear model with no optimization) obscures the value of ML methods proposed in the research literature.
Using these insights, we propose some best practices that will enable practitioners to more effectively study and deploy ML models in clinical settings.
\end{abstract}

\section{Introduction}

Machine Learning (ML) is an increasingly powerful tool which has already delivered benefit to organizations in both the public and private sectors.
Given a set of historical data, these models identify patterns and generate inferences for previously unseen data points.
The prevalence and success of ML models is driving an increased research focus in predictive modeling across several domains, including healthcare.
These approaches show promising performance in diagnostics, patient monitoring, prediction of postoperative complications, clinical trial matching, and numerous other applications~\cite{ai-vs-radiology-residents,multicenter-deterioration,nsqip-postop,sepsis1, covid-mortality, mi3, prism}.

Predictive modeling initially leveraged more traditional statistical methods (e.g. generalized linear models) but has since expanded to include a wide variety of algorithms with a correspondingly large range of model mechanics and complexity~\cite{ml-overview}.
Recently, more complex ensemble models and high-parameter systems such as deep neural networks have gained popularity.
These models can fit intricate patterns in the training data, which potentially improves out-of-sample accuracy. The complexity, heterogeneity, and often unstructured nature of healthcare data makes the allure of advanced model types understandably strong.
However, there are barriers to developing and deploying complex ML models in a healthcare setting.
The challenges include high startup costs~\cite{hbr-adoption,adoption-factors}, siloed data sources~\cite{nature-data-importance,adoption-factors,Habehh2021-ib,ALANAZI2022100924}, and trade-offs between model complexity and the need for transparency~\cite{xai-healthcare-unreliable, ALANAZI2022100924, jmir-adoption, Habehh2021-ib}. High-stakes fields (including healthcare) require mechanisms for auditing the inference process and need diverse datasets that represent multiple sub-groups in the broader population. Prior to implementing complex models in these settings, the performance benefit must be rigorously tested in comparison to approaches that maintain model transparency.

In this work, we propose the use of \textit{strong baseline models} as a mechanism for addressing and informing approaches to these common challenges.
We define a \textit{strong baseline} as a statistical or ML model with the following properties:
\begin{enumerate}
    \item The model inference process is directly interpretable by a practitioner, without requiring methods for post hoc interpretation (which may or may not be meaningful~\cite{shap-weakness, xai-healthcare-unreliable})
    \item The model has been sufficiently tuned/optimized by a researcher for the given task (e.g. by introducing non-linear terms or interaction effects when statistically significant)
    \item The model is evaluated with the appropriate metrics, reasonably which includes some measure related to clinical utility and the use of proper scoring rules for classification tasks.
\end{enumerate}
\noindent
In contrast, we define a \textit{weak baseline} as a model that does not meet these criteria, such as one that has undergone no optimization for the given task or has very limited explainability. It is common for healthcare ML research works to compare against a weak baseline model~\cite{covid-mortality,automl,weak-baseline-2,weak-baseline-3} or omit comparisons to strong baselines entirely~\cite{mi3, no-baseline-2, npj-covid, no-baseline-4, no-baseline-6, no-baseline-7, no-baseline-8,transformer-hf}. Compared to many proposed ML approaches, a strong baseline model can offer lower startup costs and higher interpretability. Comparing new, complex predictive models against stronger baselines can help practitioners navigate the trade-offs between cost, complexity, and transparency. \\

\noindent
\textbf{Contributions. }
We provide two main research contributions:
\begin{enumerate}
    \item \textit{Demonstration of rigorous performance assessments:} Through several case studies, we demonstrate concretely how the benefits of model complexity are conditional on the performance of strong baseline models.
    
    \item \textit{Conceptual evaluation framework:} We enumerate several key considerations for practitioners to more effectively and fairly create and evaluate models, as well as meaningfully discuss trade-offs when contributing to healthcare ML literature.
\end{enumerate}

\section{Statistical Baselines}

Healthcare ML research inconsistently employs \textit{baseline models}, which are used as a benchmark for evaluating the comparative performance of a proposed predictive model. Generalized linear models (GLMs) such as linear or logistic regression are frequently used as baseline models for comparison. Improved performance over these baselines can then be used as a justification for the increased complexity of the proposed approach or for excluding other statistical models from the evaluation. This procedure is both practical in the model selection process and consistent with foundational principles such as Occam's Razor.

Generalized linear models are parametric models that require a linearity assumption. For example, linear regression assumes that the conditional expectation of the response given the predictors is linear \textit{in the model parameters}. However, when GLMs are used as baselines in the existing literature, it is remarkably common to make the more restrictive assumption that \textit{the model must be linear in the predictor variables}~\cite{no-baseline-3,no-baseline-2,weak-baseline-2,automl}. This assumption often produces weak baseline models, which can easily be dismissed.

In the following case studies, we consider two varieties of GLMs - logistic regression (LR) and generalized additive models (GAM), as well as decision trees as candidates for strong baselines. These models have previously been applied to a wide variety of prediction tasks~\cite{lr-healthcare-applications, gam-healthcare-1, gam-healthcare-2}. Furthermore, several standard techniques exist to optimize these baseline models:

\begin{itemize}[leftmargin=*]

\item \textit{Class Balancing:} For classification tasks, particularly with significant imbalance in class frequency, we apply sample weighting inversely proportional to class frequency. We only defer to sample balancing (e.g. SMOTE~\cite{smote}) if programming APIs are unavailable for weighting.

\item \textit{Interactions:} Interaction terms may be used to fit more complex relationships to a response variable of interest. No higher than first-order interactions will be used, which maintains a high level of interpretability.

\item \textit{Non-linear Features (LR):} Datasets may be augmented to include quadratic or cubic terms, which allows LR to model non-linear relationships between the covariates and the response variable.

\item \textit{Regularization:} If the number of predictors is sufficiently large, LASSO or other regularization techniques \cite{elements-of-statistical-learning} may be applied to maintain a high degree of interpretability.

\end{itemize}

\noindent
Each case study will outline the specific techniques used for the baseline models.

\section{Case Studies}

To demonstrate the application and utility of strong baselines, as well as pitfalls that may occur when using complex models, we examine several works in the healthcare ML literature. For each study, our objective is to compare the published model to a well-constructed baseline. To be considered for this analysis, the following criteria had to be met:
\begin{enumerate}
    \item \textit{Journal:} Impact factor greater than 3.
    \item \textit{Recency:} Published within the last 5 years.
    \item \textit{Citations:} Average of 10 or more citations per year since publication.
    \item \textit{Accessibility:} Data publicly available or accessible with permission.
\end{enumerate}

\noindent
Five works were selected for further analysis (table \ref{tbl:studies}). Notably, these papers were not selected because we believe them to have any unique methodological deficiency.
Rather, they are representative of many contributions to the field in their general structure, methods, and presentation.

\begin{table*}[htbp]
    \small
    \centering
    \begin{tabular}{|p{6.5cm}|p{0.8cm}|p{2.2cm}|p{1.4cm}|}
        \hline
        \textbf{Title} & \textbf{Year} & \textbf{Journal} & \textbf{Citation Count} \\
        \hline
        Machine learning-based prediction of COVID-19 diagnosis based on symptoms~\cite{npj-covid} & 2021 & Nature Digital Medicine & 333-455 \\
        \hline
        Enhancing heart disease prediction using a self-attention-based transformer model \cite{transformer-hf}
        & 2024 &  Scientific \; \; \; Reports & 9-18 \\
        \hline
        Automated machine learning (AutoML) can predict 90-day mortality after gastrectomy for cancer~\cite{automl} & 2023 & Scientific \; \; \;  Reports & 5-10 \\
        \hline
        Individual-Level Fatality Prediction of COVID-19 Patients Using AI Methods~\cite{covid-mortality} & 2020 & Frontiers in Public Health & 36-50 \\
        \hline
        Early prediction of sepsis from clinical data: the PhysioNet/Computing in Cardiology Challenge 2019~\cite{physionet-challenge} & 2020 & Critical Care Medicine & 300-400 \\
        \hline
    \end{tabular}

    \caption{Publications Selected for Benchmarking and Reanalysis}
    \label{tbl:studies}
\end{table*}

\subsection{PCR Testing}
\label{case:covid-pcr}

\noindent
\textbf{Learning Task.}  Our first case study examines the prediction of SARS-CoV-2 infection. This is considered a binary classification of individuals as SARS-CoV-2 positive or negative (as determined by polymerase chain reaction (PCR) testing) based on symptoms and patient characteristics. \\

\noindent
\textbf{Data Description.} Data used in this study is a subset of a public dataset published by the Israeli Ministry of Health \cite{israeli-cov-data}. The data set consists of a target variable (PCR test positivity) and seven binary input features $x_{i,j} \in \{0,1\}$ 
used for model development. These included sex, age greater than or equal to 60, known contact with a COVID-19 positive individual, presence of cough, fever, sore throat, shortness of breath, and headache.
The referenced study used data from March 22th, 2020 through March 31st, 2020 as training data and data from the subsequent week as testing data, with both sets consisting of approximately 50,000 rows.  \\

\noindent
\textbf{Previously Published Results.} An ensemble approach was applied to this dataset, using a gradient-boosted decision tree model, which is considered a state-of-the-art approach for tabular data \cite{lgbm}. This achieved impressive results (AU-ROC 0.90, AU-PRC 0.66). No comparisons were made against baseline models. SHAP values were used to estimate the impact of each feature on final predictions. \\

\noindent
\textbf{Secondary Evaluation.} We trained a gradient-boosted model was created as described in the published analysis using the same subset of data, and similar metrics were achieved (AU-ROC 0.91, AU-PRC 0.72). A logistic regression model including quadratic terms and balanced class weighting was fit as well, and performance was compared to the gradient-boosting approach using accuracy, sensitivity, specificity, F1 score, AU-ROC, and AU-PRC.

\begin{figure}
    \centering
    \includegraphics[width=0.70\linewidth]{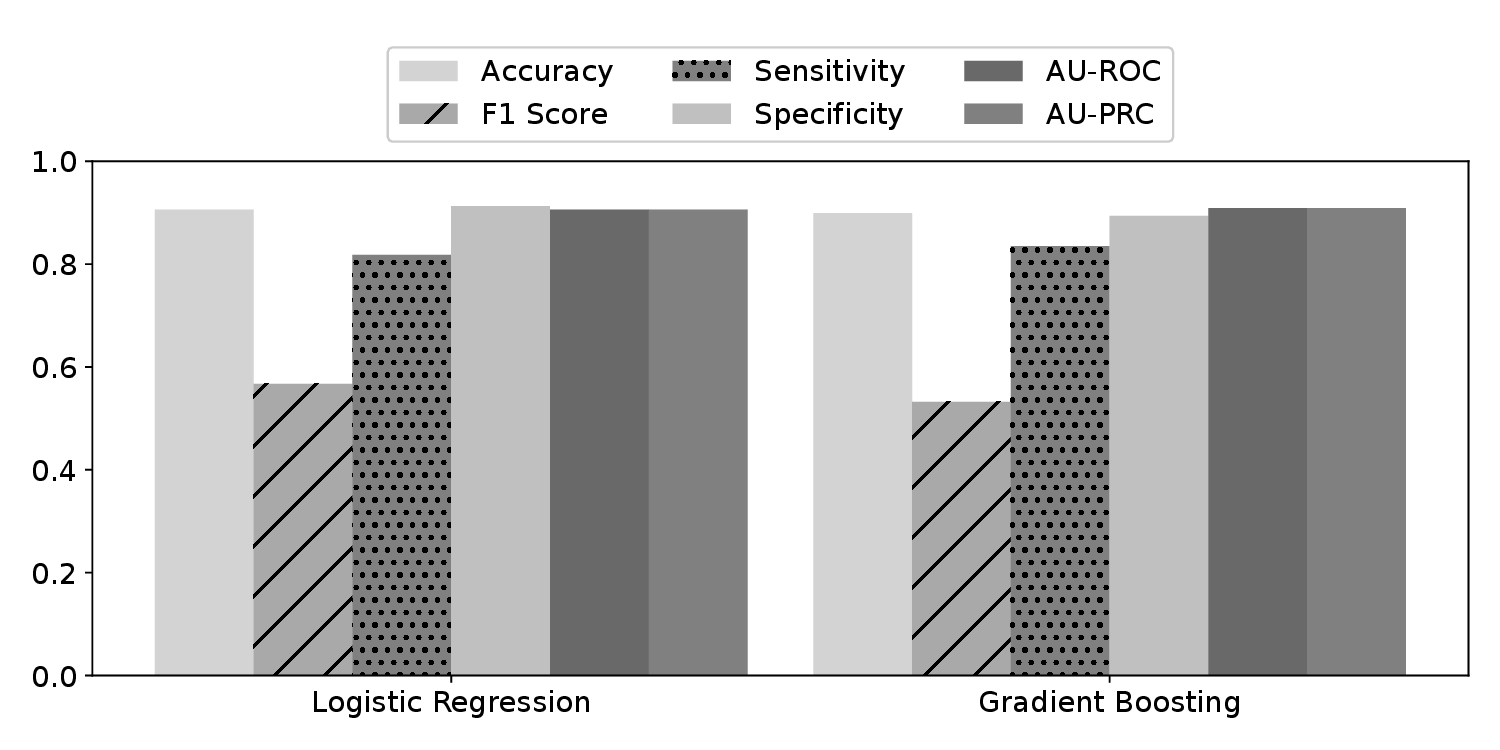}
    \caption{Model comparison for SARS-CoV-2 PCR testing}
    \label{fig:covid-pcr-model}
\end{figure}

Figure \ref{fig:covid-pcr-model} shows the results for this experiment. Overall, performance was extremely similar between the two approaches, with LR demonstrating marginally superior accuracy, F1, and specificity and the gradient boosting model marginally superior sensitivity, AU-ROC, and AU-PRC. \\

\noindent
\textbf{Key Takeaway} 
This finding reflects the need for systematic evaluation against baseline models. Assuming that a particular model type is optimal may prevent the discovery that a simpler or more interpretable model may be sufficient for a given task.

\subsection{Heart Disease Prediction}
\label{case:heart-disease}

\noindent
\textbf{Learning Task.}  The next case study explores heart disease modeling. This was framed as a binary classification task, with patient risk factors, lab values and other characteristics used to predict the presence of heart disease (defined as 50\% or greater narrowing of one or more major cardiac vessels).\\

\noindent
\textbf{Data Description.} Data was sourced from the public Cleveland Heart Disease dataset from the University of California Irvine (UCI) machine learning repository \cite{uci-heart-disease}, which has been the subject of a large number of studies \cite{hf-1, hf-2, hf-3}, often involving novel ML techniques to demonstrate marginal gains in predictive accuracy. This dataset is 303 rows in length, consisting of both categorical and numeric features. These include age, sex, serum cholesterol, electrocardiography findings, and others. An expanded version of this dataset is available through Kaggle \cite{kaggle-heart-disease} with an identical structure that contains data from the Cleveland set as well as 3 additional sites totaling 1025 rows.\\

\noindent
\textbf{Previously Published Results.} The published work proposes a self-attention-based transformer model and benchmarked against recurrent (RNN), convolutional (CNN), and other neural networks. Of the Cleveland dataset, 80\% of data was used for training, 10\% for validation, and 10\% for testing and metric generation. Predictive accuracy was used as the primary comparison between models. The proposed approach was found to outperform other neural network approaches as well as numerous prior modeling works in the literature, achieving an accuracy of 96\%.  \\

\begin{figure}
    \centering
    \includegraphics[width=0.70\linewidth]{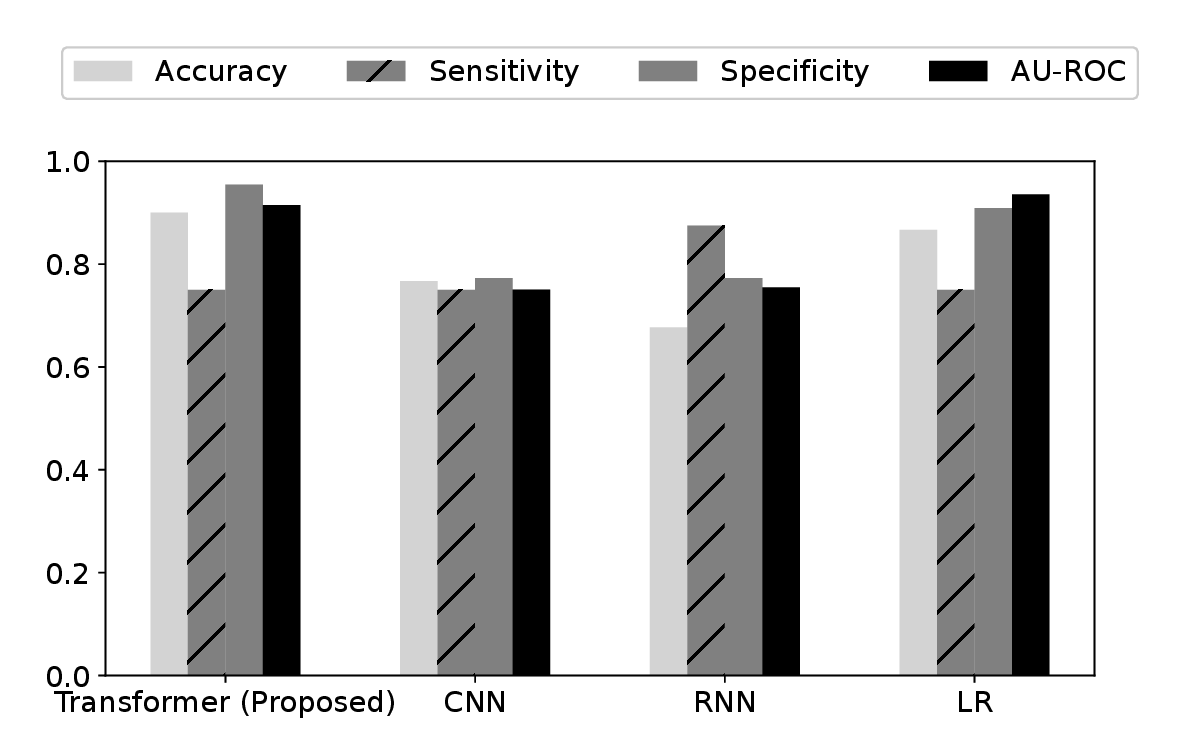}
    \caption{Model performance on Cleveland heart disease data. The transformer-based approach demonstrates superior performance on most metrics.}
    \label{fig:heart_insample}
\end{figure}

\noindent
\textbf{Secondary Evaluation.} The dataset was standardized to an approximately normal distribution, and a transformer-based approach was created following the published methods as closely as possible. This demonstrated an accuracy of 90\% on the test set and, as demonstrated in the original work, outperformed RNN and CNN models across most metrics. An LR model was also trained found to outperform RNN and CNN, showing similar performance to the proposed architecture across several metrics (Figure \ref{fig:heart_insample}), though slightly lower accuracy (90\% vs 87\%) and specificity (0.95 vs 0.91).

\begin{figure}
    \centering
    \includegraphics[width=0.70\linewidth]{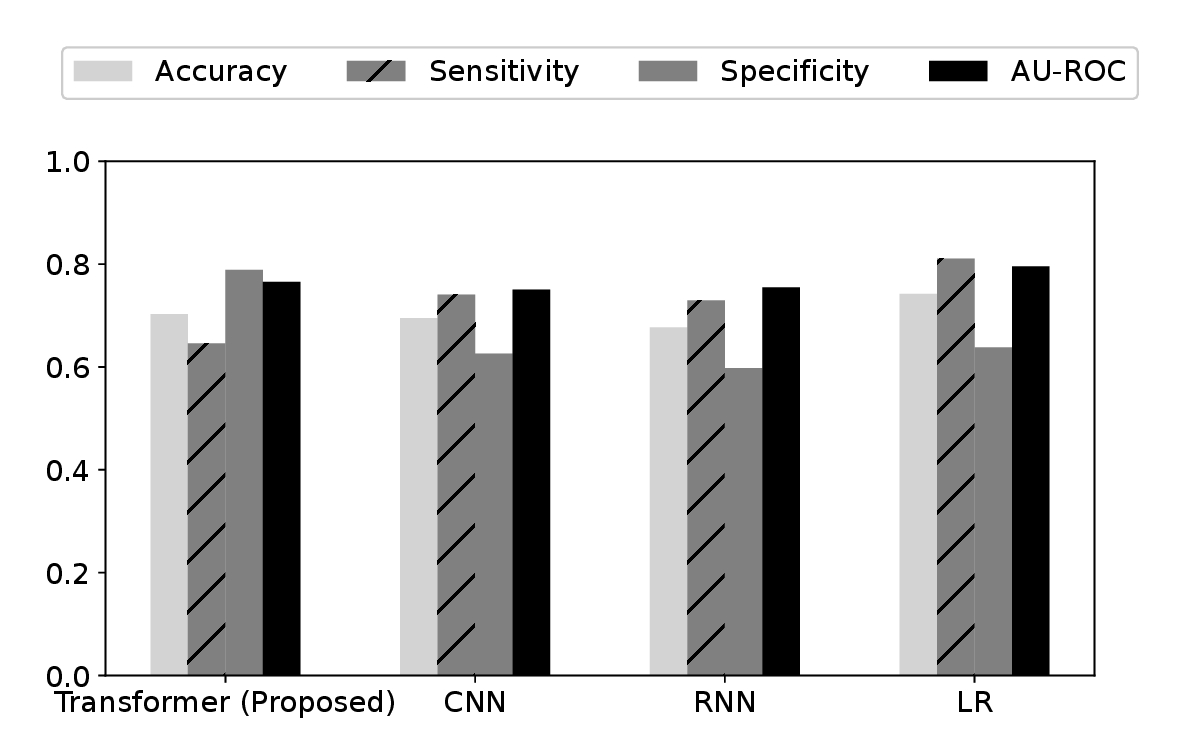}
    \caption{Model performance on data external to Cleveland dataset. Performance is similar across model types, with LR attaining the highest AU-ROC, sensitivity, and accuracy, and the transformer model the highest specificity.}
    \label{fig:heart_outsample}
\end{figure}

Testing of the same set of models was then performed using out-of-sample data from non-Cleveland sites. K-nearest-neighbor imputation was used for missing data with $k=5$. As shown in Figure \ref{fig:heart_outsample}, not only did the transformer-based model performance decline, out-of-sample performance was \textit{nearly equivalent} to the more conventional neural network architectures in accuracy and AU-ROC, with the transformer model attaining higher sensitivity and CNN and RNN models higher specificity. Furthermore, the LR model displayed \textit{higher} accuracy than the proposed approach (0.74 vs 0.70), sensitivity (0.81 vs 0.65), and AUC (0.80 vs 0.77) at the cost of some specificity (0.64 vs 0.79). \\

\noindent
\textbf{Key Takeaway} 
There are risks when fitting highly complex models to small, curated datasets. In this case, it appears our transformer-based model's performance gain largely came from more closely fitting noise or unmeasured features specific to the Cleveland dataset as it failed to generalize when compared to other approaches on the expanded dataset.

\subsection{Gastrectomy Mortality}
\label{case:automl}

\begin{figure}
    \centering
    \includegraphics[width=0.70\linewidth]{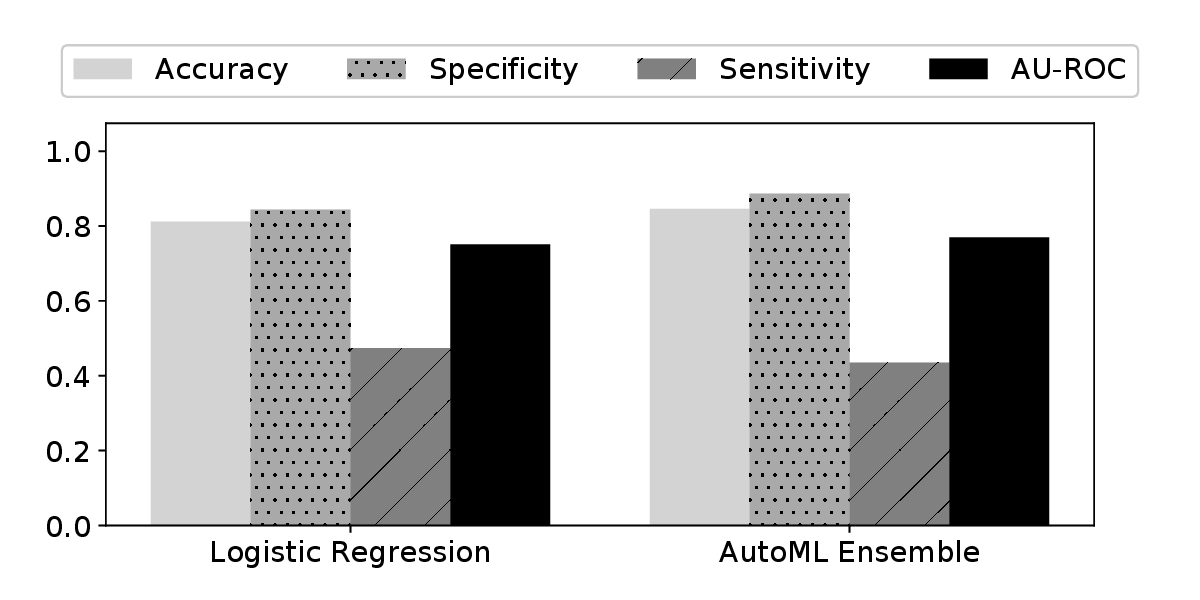}
    \caption{Model comparison for postoperative 90-day mortality}
    \label{fig:automl}
\end{figure}

\noindent
\textbf{Learning Task.} In our next case study, we examine methods for predicting the risk of postoperative mortality for patients receiving surgery for gastric cancer.
Prior works formulate this as a binary classification task, where the model predicts whether or not a patient will die within 90 days of the operation. \\

\noindent
\textbf{Data Description.} This study uses data from the National Cancer Database, covering patients who underwent gastrectomy between 2004 and 2016~\cite{automl}. The response variable is encoded as $y_i \in \{ 0, 1 \}$, where 1 indicates the specified patient died within 90 days of the operation. Twenty-six covariates are specified in this dataset, and include age, sex, tumor size, Charlson-Deyo score, prior use of radiation/chemotherapy, among other things. This dataset exhibits class imbalance, where the number of deaths accounts for only 8.8\% of the data. \\

\noindent
\textbf{Previously Published Results.} The existing work leverages AutoML as a mechanism for selecting a model with minimal oversight~\cite{automl}. Only the best model was reported in the results, which was a stacked ensemble model. This proposed approach attained an accuracy of just over 84\% with an AU-ROC around 0.77. \\

\noindent
\textbf{Secondary Evaluation.} We mirror the methodology used to evaluate the ensemble model in our own evaluation. We use 5-fold cross validation to evaluate the proposed baseline, with missing values imputed using K-nearest-neighbors with $k=5$.
Figure \ref{fig:automl} shows the resulting comparison for our logistic regression model and the published ensemble model. The LR model attains a higher sensitivity (0.47 vs 0.43) while the ensemble model performs better in the remaining categories (e.g. 0.75 vs 0.77 for AUC and 81\% vs 84\% accuracy). \\

\noindent
\textbf{Key Takeaway} Even if researchers compare against reasonable baselines, reporting these results is an important prerequisite for understanding research contributions and the relative benefit of model complexity. In this case, while the ensemble model attaining marginally better performance by some metrics, the simpler approach may be preferred in practice for interpretability, ease of implementation, or other factors.

\subsection{SARS-CoV-2 Mortality}
\label{case:covid}

\noindent
\textbf{Learning Task.} We now examine a method for identifying patients with an elevated risk of case fatality after testing positive for SARS-CoV-2~\cite{covid1,covid2,covid3,covid4,covid-mortality}. 
Prior works formulate this as a binary classification task, where the model predicts whether or not a case fatality will occur given patient characteristics. \\

\noindent
\textbf{Data Description.} This study uses a snapshot from a publicly available data source and contains roughly 29,000 cases~\cite{covid-mortality, covid-mortality-data-source}. The response variable is encoded as $y_i \in \{ 0, 1 \}$, where 1 indicates a case fatality occurred for the specified patient after testing positive for SARS-CoV-2. Eleven covariates are specified in this dataset, and include age, sex, location information, history of chronic disease, among other things. Unsurprisingly, this dataset exhibits substantial class imbalance, where the number of positive outcomes accounts for only 1.83\% of the data. \\

\noindent
\textbf{Previously Published Results.} Stacked autoencoders have been proposed for predicting whether or not a patient survived after a positive test~\cite{covid-mortality}. This approach frames the learning task as an \textit{outlier detection problem}, to account for the severe class imbalance. The autoencoder is modeled exclusively on the negative cases, then any anomalous cases that deviate from the learned negative distribution are inferred as positive cases. This proposed autoencoder approach is able to delineate between positive and negative cases, attaining an accuracy of just over 90\% with an AU-ROC around 0.7. \\

\noindent
\textbf{Secondary Evaluation.} Several baseline models were considered when evaluating the autoencoder; however, \textit{class imbalance was not considered when evaluating the baselines}. To understand the effect of this decision, we fit two additional models to the data: (1) a weighted logistic regression model with class weights inversely proportional to the frequency in the training data and (2) a weighted Generalized Additive Model (GAM). In both cases, we first standardized the data to have mean zero and unit variance. The published work reported using 30\% of the data as a test set (with no mention of cross-validation). We used 3-fold cross validation, which provides us with comparable (slightly less) training data and more robust estimates of model performance.

\begin{figure}
    \centering
    \includegraphics[width=0.6\linewidth]{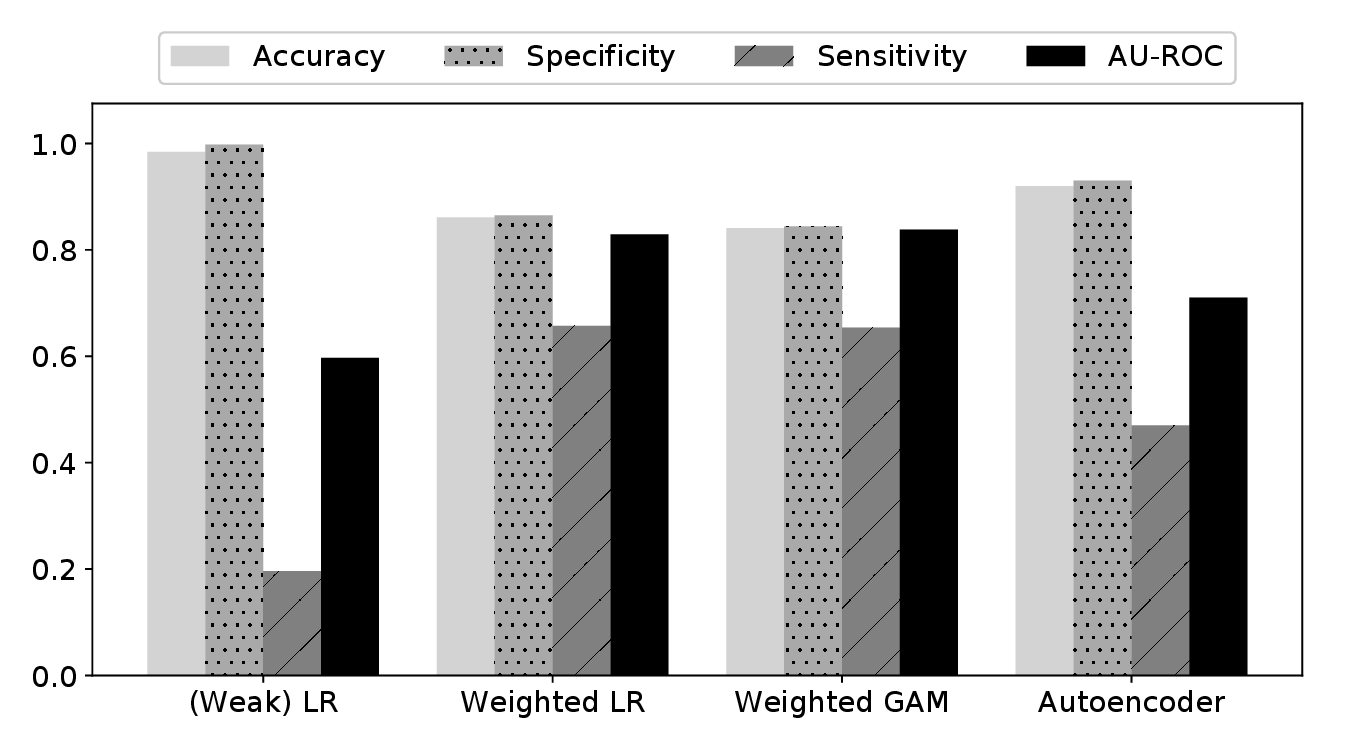}
    \caption{Model comparison for risk of SARS-CoV-2 case fatality. Weighted LR and Weighted GAM attain slightly lower accuracy, with a higher sensitivity and AU-ROC compared to the proposed autoencoder.}
    \label{fig:covid-model}
\end{figure}

Figure \ref{fig:covid-model} shows the results for this experiment. Both the weighted logistic model and the weighted GAM attain slightly lower accuracy compared to the autoencoder. However, both of our baselines attain a noticeably higher AU-ROC (about 0.83), along with a 1.5x increase in model sensitivity. \\

\noindent
\textbf{Key Takeaway}
Baseline models should be comparable to the finalized approach, ensuring a balanced comparison to the proposed deep learning methods.
In this case, practitioners may prefer to use the more transparent modeling options, given the performance characteristics.

\subsection{Sepsis Forecasting}
\label{case:sepsis}

\noindent
\textbf{Learning Task.} For this case study, we examine models for predicting if a patient is likely to develop sepsis in the near future. This is typically formulated as a binary classification task over time series data, where the response variable encodes low vs high risk of developing sepsis~\cite{sepsis1, sepsis2, sepsis3, sepsis4, sepsis5, sepsis6}. \\

\noindent
\textbf{Data Description.} In 2019, the PhysioNet/Computing in Cardiology Challenge was conducted to advance research in early sepsis detection~\cite{physionet-challenge}. The challenge provided clinical data for two healthcare facilities. The task was to use data from these two facilities (40,336 patients) to make predictions for patients in a third facility (24,819 patients), the data for which was withheld from the challenge participants. The response variable is encoded as $y_i \in \{ 0, 1 \}$, where 1 is assigned to septic patients in the relevant time windows. Forty covariates are specified in this dataset, which consist of time series EMR data.

The challenge organizers also devised a utility function, which captured the asymmetries between false positives/negatives. Positive utilities indicate the predictions were clinically useful while negative utilities encoded false negatives, i.e. failing to detect the presence of sepsis in a clinically useful time frame. \\

\noindent
\textbf{Previously Published Results.} Top-performing models in the contest utilized gradient boosting for predicting sepsis risk. However, after the submission period, the contest organizers published results that indicated the top five highest scoring teams all achieved a \textit{negative utility} on the out-of-sample data, indicating the models did not generalize well to the third healthcare facility~\cite{physionet-challenge}. Presumably, this was an unexpected result, given that the winning team achieved an AU-ROC of 0.868 using 5-fold cross-validation on the training data~\cite{sepsis-winner}. Furthermore, the competition organizers observed that traditional model metrics used to evaluate models failed to capture or correlate with clinical utility~\cite{physionet-challenge}. \\

\noindent
\textbf{Key Takeaway} 
In this case, it is evident that the ensemble models were overfitting the training data, either due to (1) insufficient regularization in the training process, (2) a low signal-to-noise ratio in the underlying data, or (3) an expectation that data drawn from a third facility would be approximately identically distributed with the data from the training facilities. In any case, reducing model complexity would likely result in more consistent model performance across facilities. We do not contribute a novel evaluation for this study, since the out-of-sample data was never made publicly available. Rather, we use the published results to call attention to the fact that stronger baselines would have been beneficial in this setting and traditional model performance metrics failed to capture clinical usefulness.

\section{Discussion}

Several existing research works do not evaluate against any baseline~\cite{mi3,npj-covid,no-baseline-4,no-baseline-7,no-baseline-8} or omit comparisons to strong baselines~\cite{no-baseline-2,no-baseline-6,transformer-hf}.
This makes it difficult to determine if there are any penalties associated with model transparency.
As we observed in the PCR testing and heart disease case studies (sections \ref{case:covid-pcr} and \ref{case:heart-disease}), complete exclusion of baselines can prevent practitioners from recognizing circumstances where increased model complexity provides no improvement in out-of-sample performance.
Similarly, many works include a weak baseline for comparison without additional optimization~\cite{covid-mortality,automl,weak-baseline-2,weak-baseline-3}.
These baselines are often limited to linear terms, ignore interaction-based effects, or fail to properly address class imbalance.
This issue was observed in the case study that modeled SARS-CoV-2 mortality (section \ref{case:covid}), where a well-constructed baseline outperformed not only the originally published baseline, but the proposed deep learning model as well.
Omitting baseline models or comparing against a simplistic model can significantly hamper the reader's ability to understand novel modeling applications in healthcare ML literature. In particular, when evaluating a novel method, a key consideration is what additional benefits are provided by the new approach over existing frameworks. However, this is difficult or impossible to assess without adequate baseline models. Benchmarking against a weak model offers readers a data point for comparison; however, it ultimately fails to capture the expected performance gains of a novel approach against conventional methods.

Exploratory or statistical analysis of healthcare datasets is often necessary to understand unique subgroups, institutional patterns, or cohorts with rare diseases.
These analyses can provide unique insights despite limited size, sampling biases, and class imbalances. However, these common features of healthcare data make predictive modeling tasks challenging, especially when applying conventional statistical approaches for the purpose of making predictions.
Without accounting for the data considerations common in these types of datasets, model performance may initially appear inadequate or sub-optimal.
This can lead to incorrectly inferring that a baseline is inferior for a task and lead to pursuing the training of more complex, high-parameter models.
As demonstrated in this study, some advanced models carry an increased risk of fitting noise or unmeasured patterns within small datasets.
This leads to poor generalizability on out-of-sample data or different populations.
An example of this issue is seen in the heart disease case study (section \ref{case:heart-disease}). Public datasets, particularly those limited in size, can result in
analyses that demonstrate incremental performance improvements through increasingly elaborate techniques ~\cite{hf-complicated-1, hf-complicated-2, hf-complicated-3, hf-complicated-4}.

A final challenge in the healthcare ML domain is the use or creation of metrics which accurately convey the utility of a model. For example, in the sepsis forecasting case study (section \ref{case:sepsis}), the authors created a novel metric to more fully capture how predictions translate to clinical benefit. Their analysis showed little relationship between clinical utility and common metrics such as AU-ROC.
Furthermore, learning tasks in the literature are often formulated as classification problems, where the model is evaluated on its ability \textit{to make binary decisions} rather than improve the decision-making process for practitioners.
Many highly successful predictive tools used
in practice inform clinicians through risk scores or forecasting which necessitate a final, binary decision made by an expert~\cite{nsqip-postop, ASCVD}.  \\

\noindent
\textbf{Limitations.}
For several case studies where code was not available for exact replication of the model used by the original study, an attempted re-creation of the proposed model was carried out according to the published methods. While we cannot claim the production of an identical model, examples were only included where metrics achieved were reasonably similar to those originally published. Furthermore, even in the absence of the exact model produced by the original work, these case studies highlight potential pitfalls in similar modeling efforts.   

While many healthcare-related learning tasks are suitable for statistical modeling and comparison, cases involving extremely high-dimension data may not be feasible for such techniques. In domains such as image analysis and natural language processing, for example, more advanced techniques which can automate learning of features and representations are more practical. \\

\noindent
\textbf{Best Practices.}
As machine learning increasingly becomes a focal point of healthcare research, efforts have been made to standardize reporting of results such as the recent TRIPOD-AI guidelines ~\cite{tripod-guidelines}. To further improve discussion and consideration of trade-offs related to the common issues demonstrated in this work, we advocate for several considerations to be consistently included in research works:
\begin{enumerate}
    \item What baseline models are considered in the evaluation? Are the baselines carefully constructed, including reasonable hyperparameters and non-linear features where appropriate?

    \item How interpretable is the proposed model? Is this acceptable for potential clinical adoption?
    
    \item How are the proposed metrics aligned with clinical utility? Are classification models evaluated with respect to their generated class probabilities or obscured by decision rules?

    \item How representative is training data of the target population? Do any subgroups warrant additional testing? Is the training data of sufficient size and heterogeneity for the learning task?
\end{enumerate}

\noindent
Answers to these questions enable practitioners to make better decisions and navigate the challenge of deploying ML models in real-world, clinical settings.

\printbibliography

\end{document}